\documentclass[runningheads]{llncs}
\usepackage[T1]{fontenc}
\usepackage{graphicx}
\usepackage{cleveref}
\usepackage{svg}
\usepackage{multicol}
\usepackage{listings}
\usepackage{xcolor}
\usepackage{booktabs}
\usepackage{array}
\usepackage{caption}
\usepackage{subcaption}
\usepackage{eso-pic}
\definecolor{lightgray}{gray}{0.5}
\usepackage{eso-pic}
\definecolor{lightgray}{gray}{0.5}

\usepackage[colorinlistoftodos,prependcaption,textsize=tiny]{todonotes}

\definecolor{codegray}{rgb}{0.5,0.5,0.5}
\definecolor{codepurple}{rgb}{0.58,0,0.82}
\definecolor{backcolour}{rgb}{0.95,0.95,0.92}

\lstdefinestyle{mystyle}{
    backgroundcolor=\color{backcolour},   
    commentstyle=\color{codegray},
    keywordstyle=\color{magenta},
    numberstyle=\tiny\color{codegray},
    stringstyle=\color{codepurple},
    basicstyle=\ttfamily\scriptsize,
    breakatwhitespace=false,         
    breaklines=true,                 
    captionpos=b,                    
    keepspaces=true,                 
    numbers=left,                    
    numbersep=5pt,                  
    showspaces=false,                
    showstringspaces=false,
    showtabs=false,                  
    tabsize=2
}

\begin{document}

\AddToShipoutPictureBG*{
  \AtPageUpperLeft{
    \raisebox{-1.5cm}{
      \makebox[\paperwidth]{
        \begin{minipage}{0.9\paperwidth}
          \centering\footnotesize
          \textcolor{lightgray}{This is an open-access, author-archived version of a manuscript published in European Conference on Multi-Agent Systems 2024.}
        \end{minipage}
      }
    }
  }
}
\AddToShipoutPictureBG*{
  \AtPageLowerLeft{
    \raisebox{1cm}{
      \makebox[\paperwidth]{
        \begin{minipage}{0.9\paperwidth}
          \centering\footnotesize
          \textcolor{lightgray}{©2024 Authors \& Springer. This is the author's version of the work. It is posted here for your personal use. Not for redistribution. \\
          The definitive Version of Record is published in the Proceedings of the European Conference on Multi-Agent Systems (EUMAS 2024).}
        \end{minipage}
      }
    }
  }
}

\title{Space Debris Removal using Nano-Satellites controlled by Low-Power Autonomous Agents}
\titlerunning{Space Debris Removal with Low-Power Autonomous Agents}
\author{Dennis Christmann\inst{1}\orcidID{0009-0007-6038-1510} \and
Juan F. Gutierrez \inst{1}\orcidID{0000-0001-8509-8075} \and
Sthiti Padhi \inst{1}\orcidID{0009-0005-9453-4488} \and
Patrick Plörer \inst{2}\orcidID{0009-0005-0800-5302} \and
Aditya Takur \inst{2}\orcidID{0000-0002-6166-1449} \and
Simona Silvestri\inst{2}\orcidID{0009-0008-1440-8994} \and
Andres Gomez\inst{1}\orcidID{0000-0002-5825-3567}}
\authorrunning{D. Christmann et al.}
\institute{Institut für Datentechnik, TU Braunschweig, Germany \\
\email{\{d.christmann, gutierrezgomez, s.padhi, gomez\}@ida.ing.tu-bs.de} \and
Institut für Raumfahrtsysteme, TU Braunschweig, Germany
\email{\{firstname.lastname\}@tu-braunschweig.de}
}
\maketitle              
\begin{abstract}
Space debris is an ever-increasing problem in space travel.
There are already many old, no longer functional spacecraft and debris orbiting the earth, which endanger both the safe operation of satellites and space travel.
Small nano-satellite swarms can address this problem by  autonomously de-orbiting debris safely into the Earth's atmosphere.
This work builds on the recent advances of autonomous agents deployed in resource-constrained platforms and shows a first simplified approach how such intelligent and autonomous nano-satellite swarms can be realized.
We implement our autonomous agent software on wireless microcontrollers and perform experiments on a specialized test-bed to show the feasibility and overall energy efficiency of our approach.

\end{abstract}
\section{Introduction}

Internet of Things (IoT) devices typically consist of ultra-low-power processors with limited sensing, processing and communication capabilities. 
Significant research effort is being devoted to continuously improve both their energy efficiency and autonomy, both from the hardware and software perspectives. 
Recent developments in single autonomous, Belief-Desire-Intention (BDI), agents have enabled their deployment in tiny microcontrollers \cite{william2022increasing}, introducing a new level of local decision-making autonomy to IoT devices. 
Further works have even demonstrated the viability of building energy-efficient multiagent systems by using finely-tuned wireless protocols \cite{vachtsevanou2023embedding}.

The ability to develop highly energy-efficient multiagent systems using commercial off-the-shelf microcontrollers has many potential applications in different fields, including IoT for space.
In this demonstration, we present our recent developments in low-power autonomous agents for space debris removal. 
Current studies indicate the amount of space debris, e.g. decommissioned satellites or spent rocket stages, is increasing to the degree that specialized deorbiting missions will become a necessity \cite{braun2013active}.
Depending on the size of these debris objects, a swarm of smaller satellites might be necessary to detumble and safely lower the object. 
We have designed a multiagent system composed of two BDI agents who communicate using OpenThread, a low-power wireless communication protocol, to synchronously push a space debris object.
We implemented our autonomous agents on a low-power wireless microcontroller, housed inside a nano-satellite mock-up, called a Free-Flyer (FF).
For our experiments, the FF was placed on a specialized air-bearing test-bed for autonomous maneuvers.

\begin{figure}[b]
    \centering
    \includegraphics[height=3cm]{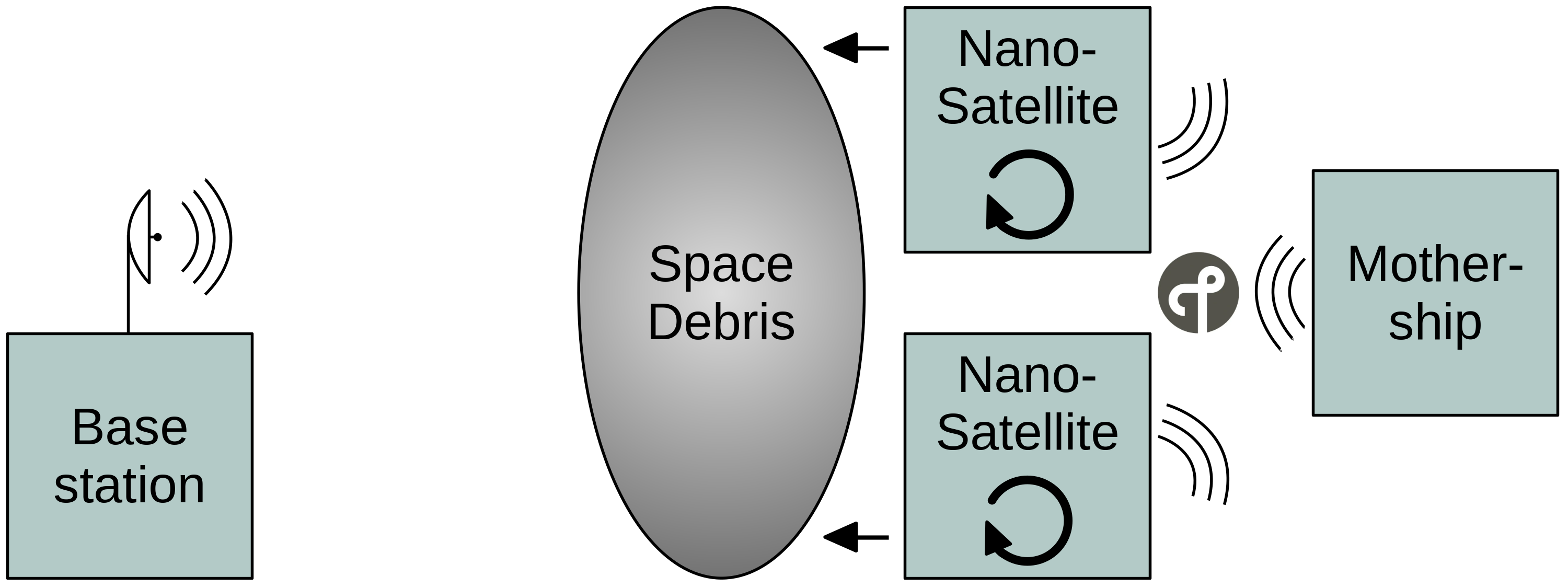}
    \caption{Simplified space-debris removal application. A base station broadcasts a mission to the mothership server. The two low-power autonomous agents synchronize via with server to deorbit debris by pushing it synchronously.}
    \label{fig:introduction}
\end{figure}

\section{System Overview}

\begin{figure}[!h]
     \centering
     \hfill
     \begin{subfigure}[b]{0.45\textwidth}
         \centering
         \includegraphics[width=\textwidth]{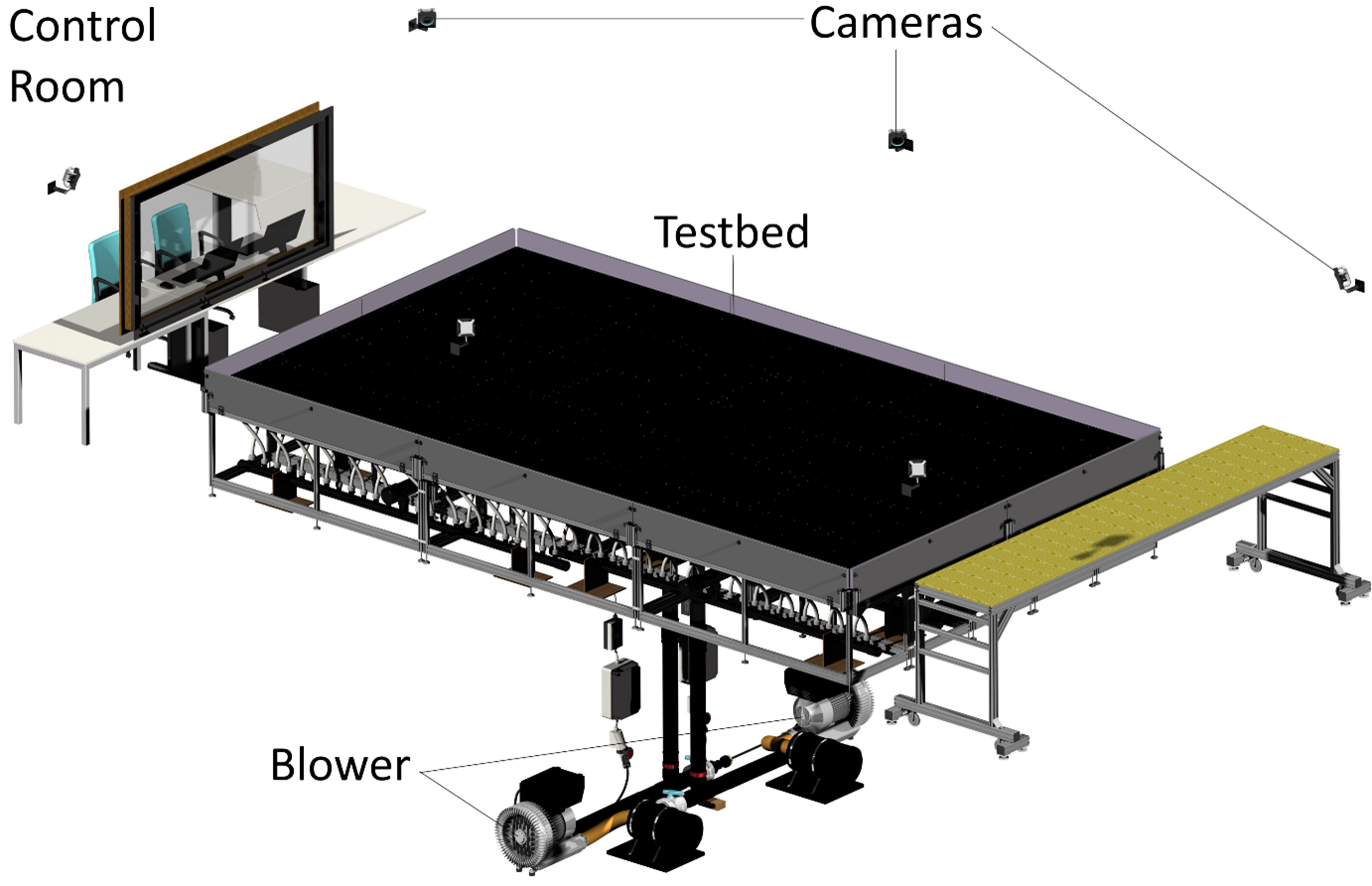}
         \caption{Overall testbed layout.}
         \label{fig:elissa:overall-layout}
     \end{subfigure}
     \hfill
     \begin{subfigure}[b]{0.3\textwidth}
         \centering
         \includegraphics[width=\textwidth]{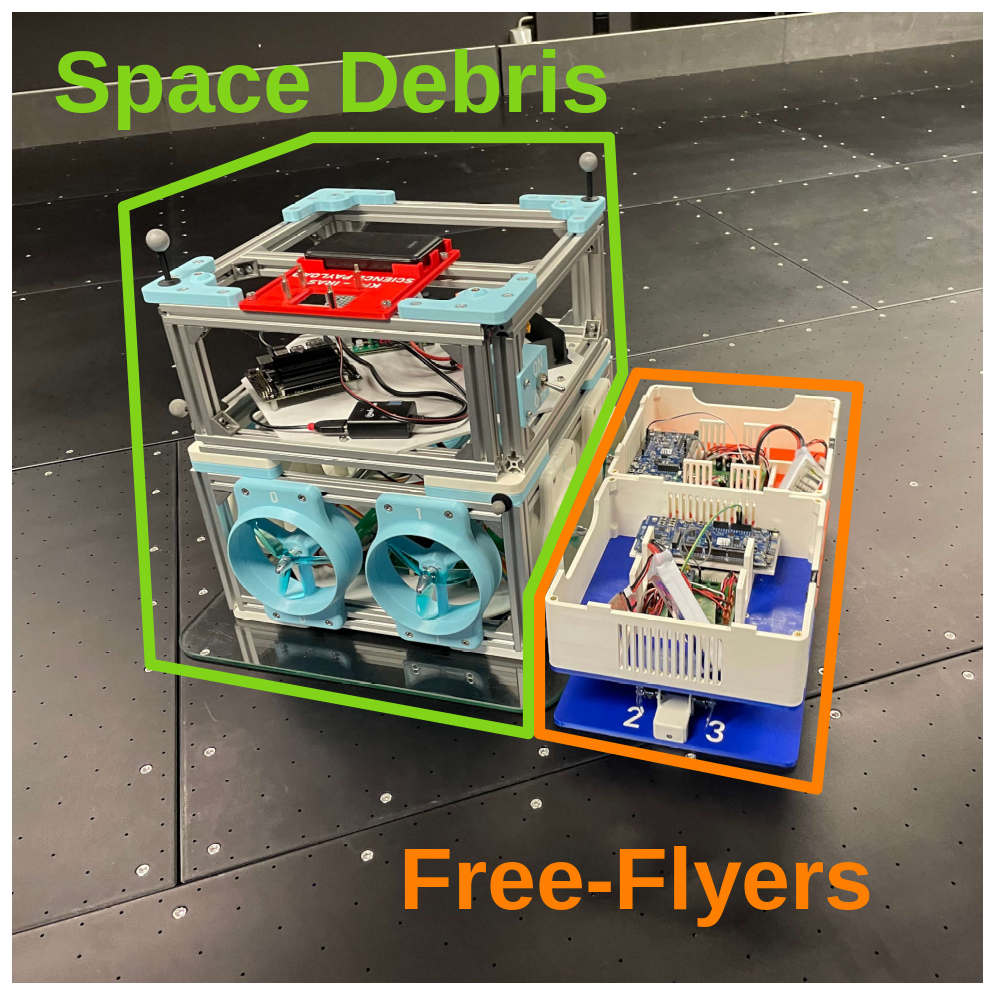}
         \caption{Two FFs push space debris away.}
         \label{fig:elissa:air-bearing-table}
     \end{subfigure}
     \hfill
        \caption{The ELISSA test-bed allows satellite mockups, called Free-Flyers (FF), to float on a cushion of air. Our autonomous agents perform the maneuvers.}
        \label{fig:elissa}
\end{figure}

Our system consists of two FFs and another nano-satellite mock-up acting as space debris (\Cref{fig:elissa:air-bearing-table}) gliding on the Experimental Lab for Proximity Operations and Space Situational Awareness (ELISSA) \cite{yang2021concept} test-bed.
In addition, an nF52840-DK is used as the mothership server and an nRF52840 dongle plays the role of a base station, completing the network shown in \Cref{fig:introduction}.

The ELISSA test-bed enables the emulation of weightlessness and contact dynamics of spacecraft in orbit using an active air-bearing table. 
The test-bed, shown in Figure~\ref{fig:elissa:overall-layout}, is mainly composed of four parts: the aeromechanical subsystem, the motion capture subsystem, the mission control subsystem and the FFs. 

\begin{figure}[!h]
     \centering
     \begin{subfigure}[b]{0.3\textwidth}
         \centering
         \includegraphics[width=\textwidth]{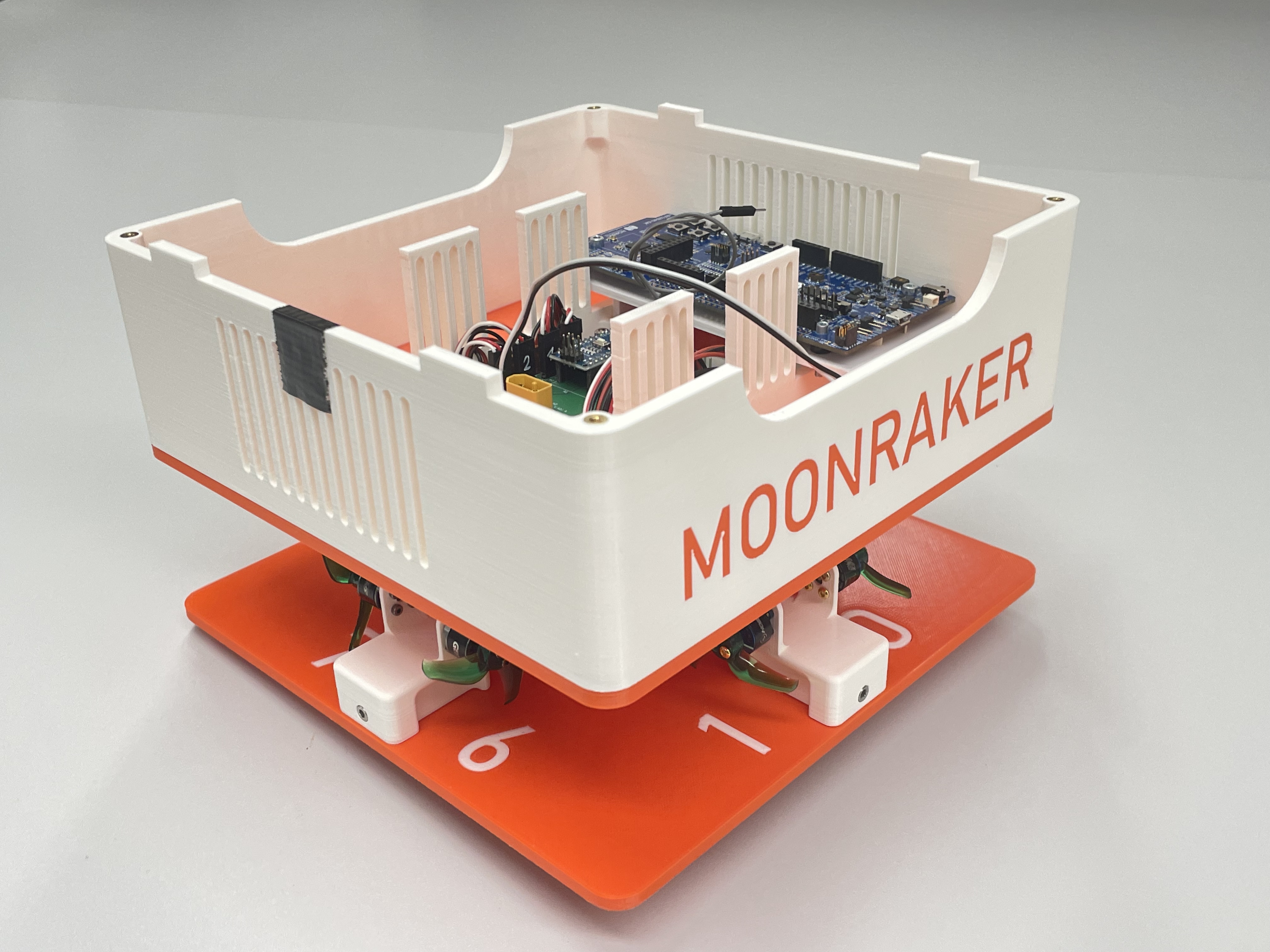}
         \caption{The FF houses the hardware.}
         \label{fig:free-flyer:real-model}
     \end{subfigure}
     \begin{subfigure}[b]{0.45\textwidth}
         \centering
         \includegraphics[width=\textwidth]{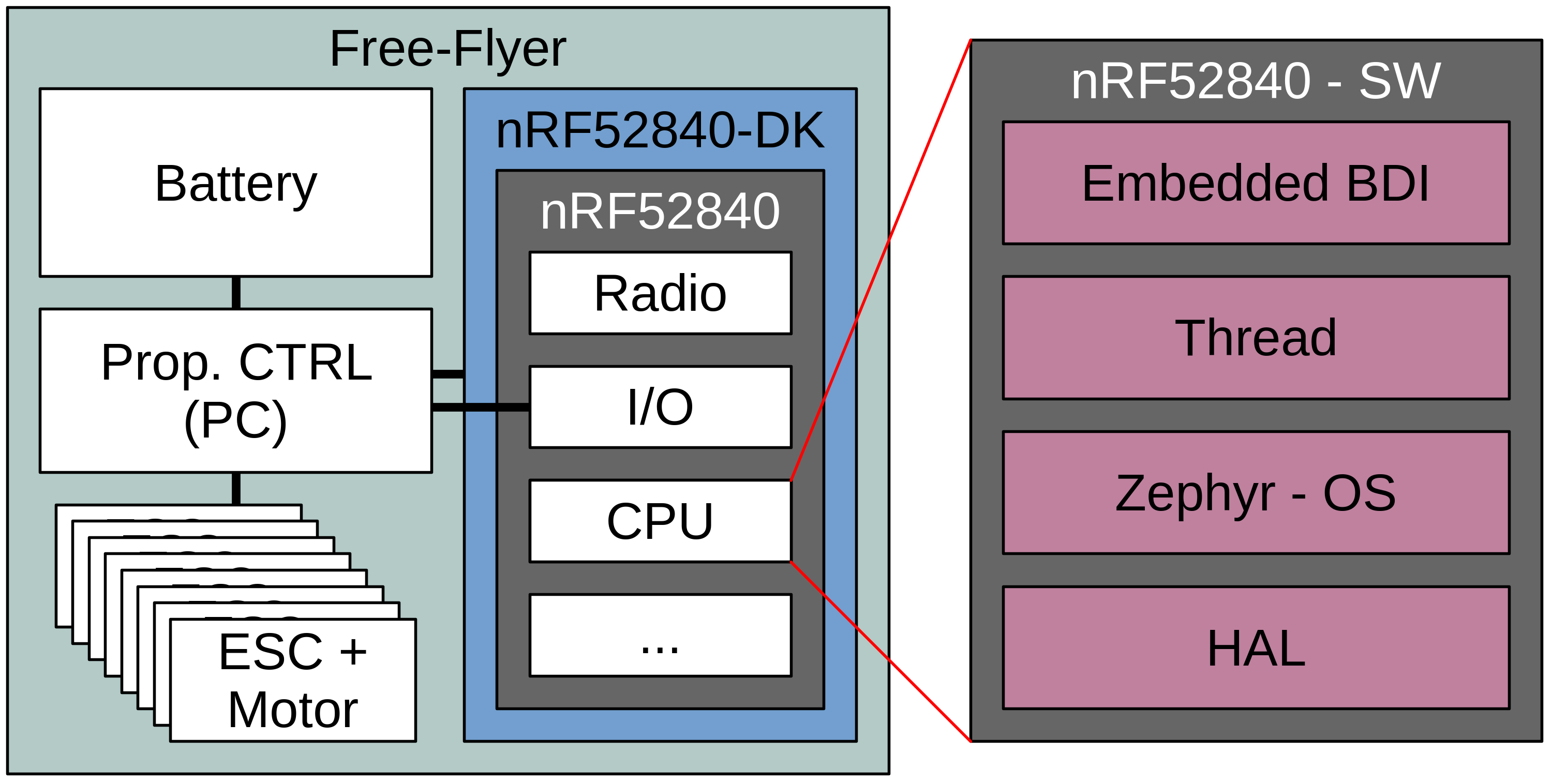}
         \caption{HW/SW architecture}
         \label{fig:free-flyer:hw-sw-diagram}
     \end{subfigure}
        \caption{The Free-Flyer (FF) floats on the ELISSA testbed. The software onboard implements the autonomous agents, which communicate wirelessly.}
        \label{fig:free-flyer}
\end{figure}

The aeromechanical subsystem transports air mass to generate an air cushion on the air-bearing table, enabling near friction-less motion for the FFs in three degrees of freedom.
The motion capture subsystems provides high-accuracy three-dimensional localization data to the (centralized) mission control subsystem, which can send wireless commands to the FFs.
The FFs have a 3D-printed main structure of the nano-sattelite mock-up, as shown in Figure~\ref{fig:free-flyer:real-model}.
It has a dimension of 22 cm × 22 cm × 26 cm and is used to mount 8 motors with propellers, various printed circuit boards (PCBs) and a LiPo battery. An additional structure on top of the mock-up allows attaching various payloads.

In this work, which focuses on autonomy for resource-constrained devices, we do not use the motion capture nor the mission control subsystems.
Instead, the reasoning and decision-making are taken by the autonomous agents running on low-power microcontrollers mounted on the FF. 

\Cref{fig:free-flyer:hw-sw-diagram} on the left shows the relevant hardware components of a FF in a block diagram.
It is moved by eight motors to which propellers are attached.
The motors are each powered by an electronic speed controller (ESC), which is controlled and supplied by the propulsion controller (PC).
The battery, which powers the entire FF, is connected to the PC.
The nRF52840-DK with the nRF52840 microcontroller on it is also connected to the PC and is responsible for autonomous and intelligent control.
The nRF52840 microcontroller sends motion commands to the PC, communicates with the CoAP server and contains an embedded BDI agent for autonomous interaction\footnote{Code and video available here: \\ \url{https://git.rz.tu-bs.de/ida/rosy/public/publication-repos/eumas-2024-demo-paper}}.

The right-hand side of \Cref{fig:free-flyer:hw-sw-diagram} shows the relevant software components required for autonomous and intelligent control.
We utilize the Zephyr real-time operating system, which provides many relevant features such as low-power operation, the OpenThread wireless protocol, multithreading, etc.
The wireless communication between the low-power autonomous agents takes place indirectly through the mothership's CoAP server. 
The Embedded BDI framework \cite{santos2022programaccao} runs as a Zephyr application, interpreting AgentSpeak code (see Listings \ref{lst:master} and \ref{lst:slave}), making use of the CoAP library to receive and send information.

\begin{center}
    \begin{minipage}[t]{0.45\linewidth}
\begin{lstlisting}[numbers=left, numberstyle=\tiny, frame=single, aboveskip=0.5cm, xleftmargin=0.5cm, caption={Master Agent Code}, captionpos=b, label={lst:master}, language={Prolog}]
!mission_search.

+!mission_search 
<- listen_gs; !!achieve.

+!achieve : new_mission 
<- announce_perform_mission. 

+!achieve : no_mission 
<- !!mission_search.
\end{lstlisting}
    \end{minipage}
    \qquad
    \begin{minipage}[t]{0.45\linewidth}
\begin{lstlisting}[numbers=left, numberstyle=\tiny, frame=single, aboveskip=0.5cm, xleftmargin=0.5cm, caption={Slave Agent Code}, captionpos=b, label={lst:slave}, language={Prolog}]
!mission_search.

+!mission_search 
<- listen_server; !!achieve.

+!achieve : new_mission 
<- perform_mission. 

+!achieve : no_mission 
<- !!mission_search.
\end{lstlisting}
    \end{minipage}
\end{center}

\begin{figure}[h]
    \centering
    \includegraphics[width=0.8\textwidth]{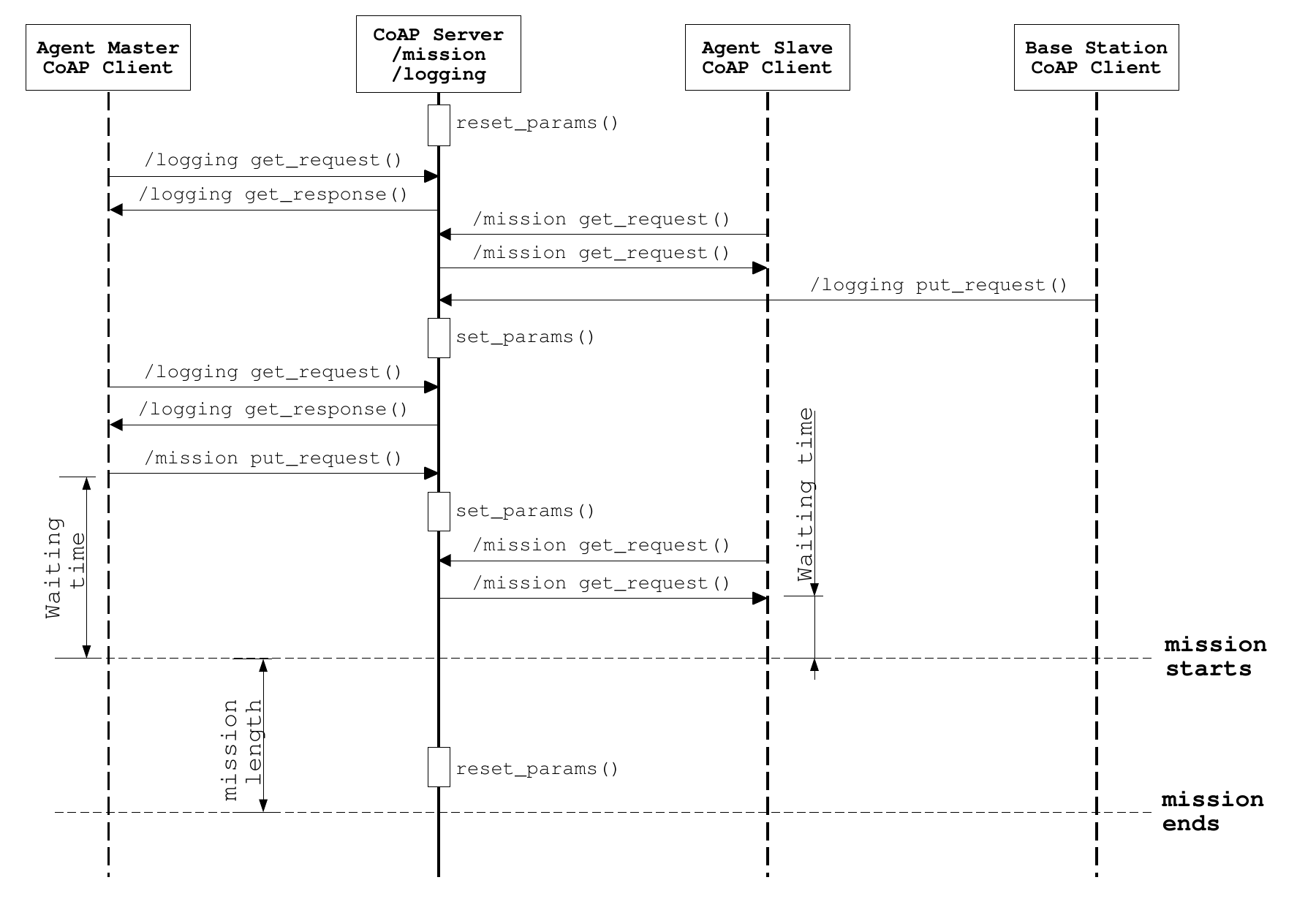}
    \caption{Timing diagram for agent-to-agent synchronization using CoAP. Base station reports a mission to the mothership, which acts as a server. Once the master agent receives the mission, the agent schedules it for the future and sends a message to slave agent with the mission information.}
    \label{fig:coap-diagram}
\end{figure}

The network, utilizing the Thread routing protocol and the CoAP application layer protocol, consists of a server and three clients (master, slave, base station). 
Two server resources, /logging and /mission, manage FF's motor speeds and the mission length. 
The base station registers these parameters with a PUT request to the server, which the master client retrieves via GET. 
The master client registers the mission start time with /mission, and the slave client polls for these parameters. 
The server synchronizes the mission despite differing wait times for master and slave clients and resets parameters post-mission for repetition.  

\section{Demonstration}

\begin{figure}[h]
    \centering
    \includegraphics[width=\textwidth]{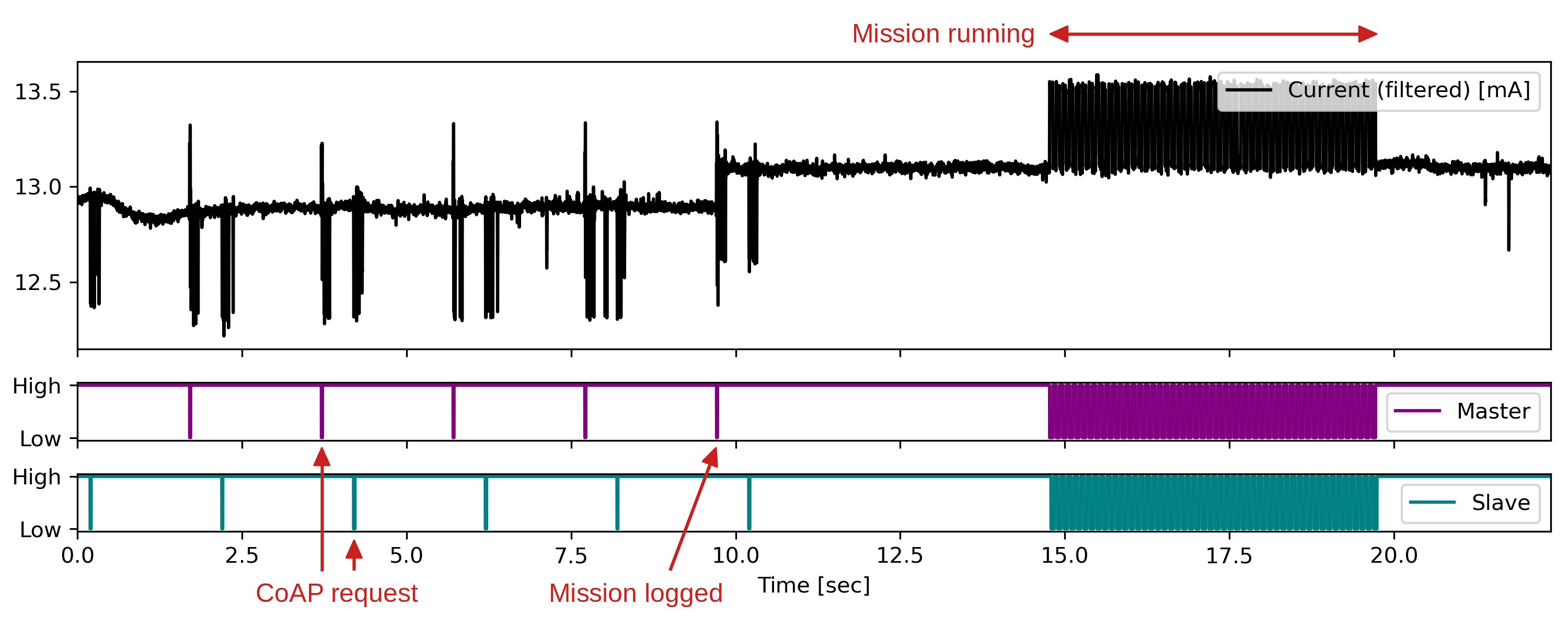}
    \caption{Filtered current measurement from master agent before, during and after mission. Lower plots show CoAP requests and the synchronized agent action.}
    \label{fig:power-and-activity-plot}
\end{figure}

We characterized the energy consumption of the agent's actions, i.e. CoAP requests, using the NRF Power Profiler.
Each action was repeated at least 210 times. 
The first action performed by the master agent, \emph{listen\_gs}, involves a 70 byte GET request which consumes on average 271\,$\mu$J and takes 14.7\,ms to complete.
The \emph{announce\_perform\_mission} action performed by the master agent, includes a 78 byte PUT request which consumes on average 294\,$\mu$J and lasts 14.7\,ms. 
Similarly, the slave agent's action, \emph{listen\_server}, involves a 70 byte GET request which consumes 276\,$\mu$J and lasts 14.9\,ms.
The synchronization error for the agents' mission was measured to be approximately 21.2 ms, derived from the timing diagram presented in Figure \ref{fig:power-and-activity-plot}. 
The current plot depicts the consumption of the master agent and has been filtered using a digital 4-th order butterworth low-pass filter with a sampling frequency of 500 Hz. 
The \emph{mission} is represented here by the blinking of an LED on the agents' boards.

This work demonstrates the operation of a multi-agent system for space debris removal applications. 
The agents can communicate efficiently using the CoAP/Thread wireless protocols.
Future work will expand the agent's capabilities, including localization awareness and refined motor control for more complex maneuvers.
\bibliographystyle{splncs04}
\bibliography{02-refs}

\end{document}